\def\year{2022}\relax
\documentclass[letterpaper]{article} 
\usepackage{aaai22}  
\usepackage{times}  
\usepackage{helvet}  
\usepackage{courier}  
\usepackage[hyphens]{url}  
\usepackage{graphicx} 
\usepackage{natbib}  
\usepackage{caption} 
\DeclareCaptionStyle{ruled}{labelfont=normalfont,labelsep=colon,strut=off} 
\usepackage{algorithm}
\usepackage{algorithmic}

\usepackage{newfloat}
\usepackage{listings}
\floatstyle{ruled}
\newfloat{listing}{tb}{lst}{}
\floatname{listing}{Listing}
\nocopyright
\title{Towards Communication-Efficient and Privacy-Preserving Federated Representation Learning}
\author{
    {\large Haizhou Shi\textsuperscript{\rm 2},
    Youcai Zhang\textsuperscript{\rm 1},
    Zijin Shen\textsuperscript{\rm 2},
    Siliang Tang\textsuperscript{\rm 2},
    Yaqian Li\textsuperscript{\rm 1},
    Yandong Guo\textsuperscript{\rm 1},
    Yueting Zhuang\textsuperscript{\rm 2}
    }
}
\affiliations{

    \textsuperscript{\rm 1} OPPO Research Institute,
    \textsuperscript{\rm 2} Zhejiang University\\
    \{zhangyoucai, liyaqian, guoyandong\}@oppo.com, 
    \{shihaizhou, zijinshen, siliang, yzhuang\}@zju.edu.cn
}

\usepackage{bibentry}
\usepackage{xspace}
\usepackage{booktabs}
\usepackage{amssymb}

\newcommand{\iid}{i.i.d.\xspace}
\newcommand{\ours}{FLESD\xspace}
\usepackage{color}

\usepackage[ruled,vlined,linesnumbered,algo2e]{algorithm2e}

\newlength\savewidth\newcommand\shline{\noalign{\global\savewidth\arrayrulewidth
  \global\arrayrulewidth 1pt}\hline\noalign{\global\arrayrulewidth\savewidth}}
\newcommand{\tablestyle}[2]{\setlength{\tabcolsep}{#1}\renewcommand{\arraystretch}{#2}\centering\footnotesize}

\usepackage{multirow}
\usepackage{subfigure}
\usepackage{pifont}
\usepackage{amsmath,amsfonts,amssymb,bm}

\def\Dpub{{D_{\textnormal{pub}}}}
\def\lcl{{l_{\textnormal{cl}}}}

\def\Figref#1{Figure~\ref{#1}}

\def\Secref#1{Section~\ref{#1}}

\def\eqref#1{equation~\ref{#1}}
\def\Eqref#1{Equation~\ref{#1}}

\def\Algref#1{Algorithm~\ref{#1}}

\def\Tabref#1{Table~\ref{#1}}

\def\1{\bm{1}}

\DeclareMathAlphabet{\mathsfit}{\encodingdefault}{\sfdefault}{m}{sl}
\SetMathAlphabet{\mathsfit}{bold}{\encodingdefault}{\sfdefault}{bx}{n}

\def\gD{{\mathcal{D}}}

\newcommand{\E}{\mathbb{E}}

\begin{document}

\maketitle
\let\thefootnote\relax\footnotetext{This work was done during Haizhou's internship at OPPO.}
\begin{abstract}
This paper investigates the feasibility of federated representation learning under the constraints of communication cost and privacy protection. Existing works either conduct annotation-guided local training which requires frequent communication, or aggregate the client models via weight averaging which has potential risks of privacy exposure. To tackle the above problems, we first identify that self-supervised contrastive local training is robust against the non-identically distributed data, which provides the feasibility of longer local training and thus reduces the communication cost. Then based on the aforementioned robustness, we propose a novel Federated representation Learning framework with Ensemble Similarity Distillation~(\ours) that utilizes this robustness. At each round of communication, the server first gathers a fraction of the clients' inferred similarity matrices on a public dataset. Then it ensembles the similarity matrices and train the global model via similarity distillation. We verify the effectiveness of \ours by a series of empirical experiments and show that, despite stricter constraints, it achieves comparable results under multiple settings on multiple datasets. 
\end{abstract}

\section{Introduction}
Federated Learning~(FL) seeks to collaboratively train the model distributed across a large network while keeping the clients' private data protected~\cite{konecny_federated_2016, zhao2018federated, li2019fedmd, li2020federated, lin2020ensemble, wang2020federated}. Despite that significant development has been achieved by the community towards more secure, efficient, and reliable FL systems, the real-world deployment of FL still has some primary issues to solve. For instance, annotations can not be obtained on the client side in certain scenarios, which makes currently prevalent annotation-guided frameworks inapplicable.



\begin{figure}[t]
  \vspace{-0.9em}
  \centering
  \includegraphics[width=.47\textwidth]{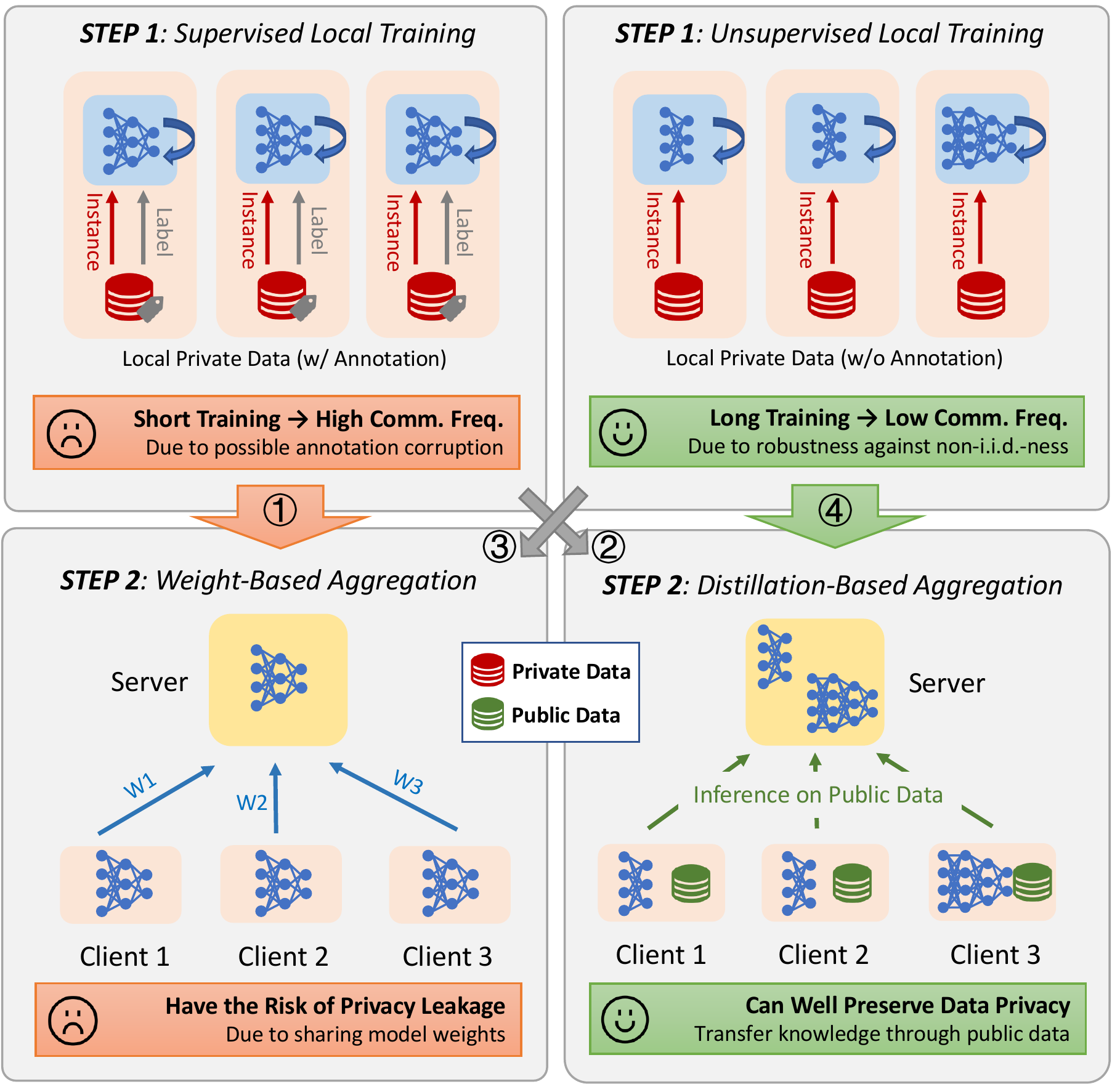}
  \vspace{-1.2em}
  \caption{Categorization of local training and global aggregation in federated learning. 
  \textbf{\ding{192}}: Classic weight-averaging supervised FL frameworks~\cite{mcmahan2017communication,li2020federated}. 
  \textbf{\ding{193}}: Distillation-based supervised FL frameworks~\cite{lin2020ensemble,gong2021ensemble}. 
  \textbf{\ding{194}}: Weight-averaging FURL frameworks~\cite{zhang2020federated,zhuang2021collaborative}. 
  \textbf{\ding{195}}: \textbf{Our proposed \ours}, first framework that possesses advantages at both local training and global aggregation, which is more communication-efficient and privacy-protective. 
  }
  \label{fig:fl-categories}
  \vspace{-0.5em}
\end{figure}

To solve this problem, the branch of Federated Unsupervised Representation Learning~(FURL) emerges~\cite{jin2020towards,van2020towards,zhang2020federated,zhuang2021collaborative}. 
It aims at training a representation encoder with the non-annotated data locally distributed across a network of clients. As a subsequent branch of general FL, existing FURL methods mostly inherit the basic setting of the previous supervised frameworks, and fail to notice the intrinsic difference between the two types of local training (discussed later in \Secref{sec:obs}). For example, FedCA~\cite{zhang2020federated} and FedU~\cite{zhuang2021collaborative} both adopt short local training time, which leads to \textit{high communication overhead}; and they both use weight-based global aggregation scheme, which is prone to potential hazard of \textit{privacy leakage}.

This paper proposes to address the aforementioned two problems in federated representation learning. 
We start off with identifying an essential feature of the prevalent contrastive representation learning: compared to the supervised methods, it's more robust against the non-identically distributed~(non-\iid) data across the clients. Therefore longer epochs of local training and lower communication frequency become viable.
Next, to leverage the robustness and provide strict constraints on privacy preservation, we propose a new framework, Federated representation Learning via Ensemble Similarity Distillation~(\ours). 
The local training of \ours is conducted self-supervisedly following SimCLR~\cite{chen2020simple}. At each communication round, the selected clients infer the similarity matrix on a public dataset and return them to the server. Then \ours transfers the knowledge from the ensembled similarity matrix to the global models and then distributes the them back to the clients for future local training.
 
From the perspective of representation learning, federated frameworks can be roughly categorized into four types based on their design of local training and global aggregation, as shown in \Figref{fig:fl-categories}. Unsupervised local training, due to its independence from the label quality, is much more robust against the non-\iid data distribution, leading to longer local training and fewer communication rounds. On the global aggregation side, distillation-based methods are designed to avoid privacy leakage and to support model heterogeneity. 

\textbf{\ours} is the first framework that belongs to the \ding{195}-th category. Comprehensive experiments show that \ours can yield good-quality representation space in a communication-efficient manner. Besides, it is less prone to privacy exposure, and has potential application to heterogeneous FL systems. To summarize, our main contributions are:
\begin{itemize}
    \item We point out that contrastive representation learning framework is robust to local training on non-\iid data distribution. Such property is exploitable in FL systems.
    \item We propose a new federated self-supervised learning framework \ours that possesses the edge of significantly fewer communication rounds during training and is applicable under strict data-privacy restrictions.
\end{itemize}

\section{Related Work}
\subsubsection{Federated Learning.}
Firstly proposed by \cite{konecny_federated_2016} for jointly 
training machine learning models across clients while preserving their data privacy, federated learning has become a key research area in distributed machine learning. One important challenge of federated learning is the non-\iid data distribution and many works have been proposed to particularly address this problem~\cite{zhao2018federated,li2020federated,li2021fedbn,wang2020federated}. 
Another focus is to solve model heterogeneity. 
\citet{smith2018federated} is the first attempt trying to apply multi-task learning framework to aggregate separated but related models; 
FedMD introduces the technique of knowledge distillation \cite{li2019fedmd,hinton2015distilling} to federated learning for client personalization;
FedDF proposes to use ensemble distillation to effectively aggregate the clients' knowledge regardless of the model architectures\cite{lin2020ensemble}. 
Different from the methods above that address federated supervised learning, \ours aims at solving the federated unsupervised representation learning in a privacy-preserving and efficient way, the main challenge of which lies in how to design an representation-level distillation mechanism that effectively aggregate the knowledge of various clients. We solve this problem by our proposed Ensemble Similarity Distillation~(ESD) technique.

\begin{figure*}[t]
    \vspace{-0em}
    \centering
    \includegraphics[width=1\textwidth]{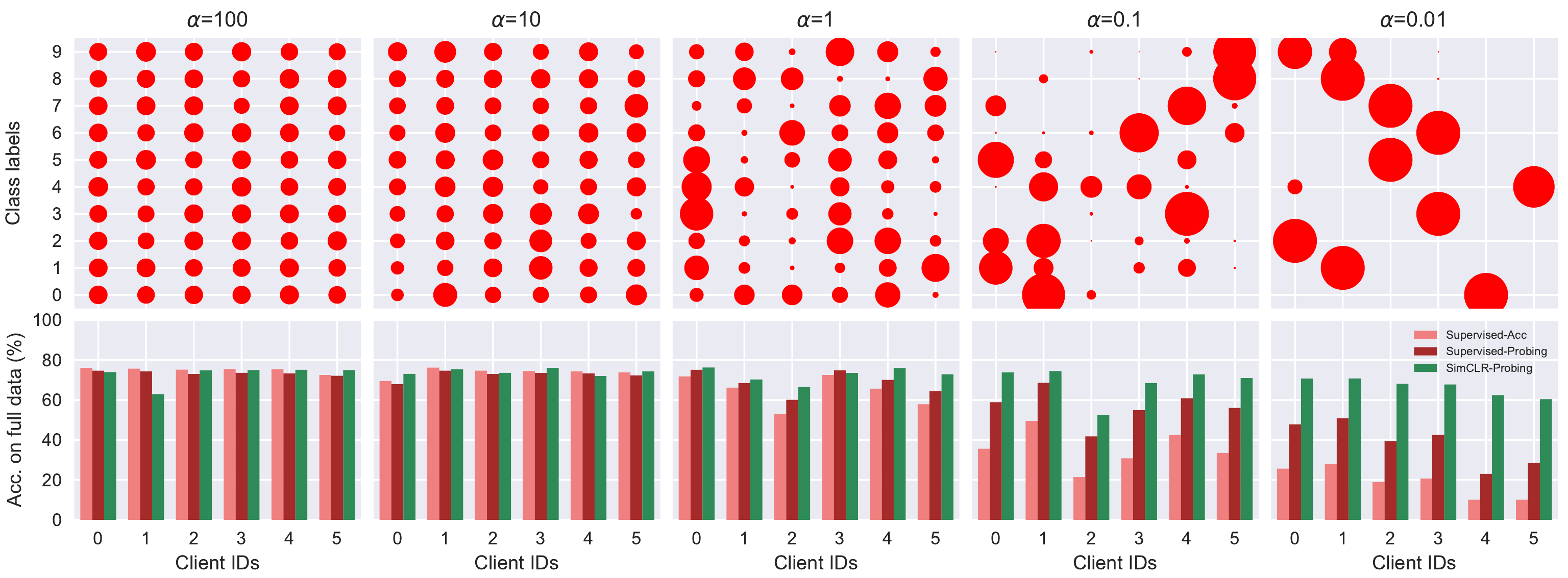}
    \vspace{-0.7em}
    \caption{Local training of supervised and self-supervised learning on CIFAR10. 
    The top row visualizes the non-\iid-ness of the data distribution; the radius of each circle represents the size of the local dataset. 
    The bottom row compares the behavior of supervised and self-supervised local training. 
    The self-supervised framework SimCLR is more robust against non-\iid-ness of the client data distribution than the supervised method. 
    }
    \label{fig:distribution}
    \vspace{0.5em}
\end{figure*}

\subsubsection{Self-Supervised Contrastive Learning.}
Contrastive learning follows the multi-view hypothesis and performs instance discrimination pretext task to train the encoder network~\cite{wu2018unsupervised, oord2018representation, tian2019contrastive, chen2020simple, he2020momentum}. 
SimCLR proposes to adopt aggressive data augmentation and large batch size to eliminate possible shortcuts in contrastive learning, yielding surprisingly good results~\cite{chen2020simple}. MoCo addresses the problem of large computational cost by introducing a momentum encoder and a momentum queue that encodes and stores the negative representations~\cite{he2020momentum,chen2020improved}.

There are some works that explore self-supervised learning in federated scenario~\cite{jin2020towards,van2020towards,zhang2020federated,zhuang2021collaborative}.
To address the problem of misaligned representation spaces yielded by client models, FedCA proposes to use a public dataset and an alignment module to constrain the local training of SimCLR algorithm~\cite{zhang2020federated}. 
FedU utilizes BYOL as its local self-supervised training framework and design a dedicated weight-averaging scheme for its predictor component, achieving the state-of-art performance on CIFAR10/100 dataset~\cite{zhuang2021collaborative,grill2020bootstrap}.
However, both methods are designed under weight-averaging aggregation framework and may cause serious privacy leakage during the actual development. They also fail to notice that the annotation-free local training objective such as contrastive representation learning is more robust against statistical heterogeneity, and therefore can be utilized towards more communication-efficient method.
To the best of our knowledge, we are the first to handle the communication efficiency and privacy issue for FURL.

\subsubsection{Knowledge Distillation.}
The technique of Knowledge Distillation~(KD) was firstly proposed to train a lightweight model under the guidance of a pre-trained high-capacity model~\cite{hinton2015distilling}. Up to today, there are generally three categories of distillation methods: response-based distillation~\cite{hinton2015distilling}, feature-based distillation~\cite{romero2014fitnets,zagoruyko2016paying,kim2018paraphrasing,heo2019knowledge,tian2019crd}, and relation-based distillation~\cite{park2019relational,lassance2020deep,koohpayegani2020compress,fang2021seed}. Our designed global aggregation is inspired by CompRess~\cite{koohpayegani2020compress} and SEED~\cite{fang2021seed}, which distill the knowledge from a pre-trained model by matching the similarity distribution. However, there are two significant differences between the previous work and our method \ours. Firstly, due to the data privacy concerns, we do not have direct access to the teacher model, and thus we have to perform the distillation on a ``off-line'' manner. Secondly, we have multiple teacher models that are trained under different local data distributions, and we devise an ensemble mechanism for the similarity matrices to form a global distillation target. 
\section{Contrastive Representation Learning: Robustness against Non-\iid Client Data}\label{sec:obs}

Although self-supervised learning has proven its effectiveness when trained with a large model on the data center~\cite{chen2020simple, he2020momentum, grill2020bootstrap, chen2020exploring, zbontar2021barlow, chen2020big, chen2021empirical, chen2020improved}, its performance and the related properties are not fully studied when the data are distributed across multiple client devices, i.e., in the federated scenario.
Here we answer the question of how self-supervised contrastive \textbf{local training} performs on the non-identically distributed client data~(also known as non-\iid-ness): what differs it from the annotation-guided training? How can we utilize its distinct properties if they exist?

\subsubsection{Setup.} 
As a classic self-supervised learning paradigm, we explore the behavior of SimCLR in the federated local training scenario and compare its differences between the traditional supervised learning~\cite{chen2020simple}. Following \citet{lin2020ensemble}, we synthesize varying degrees of non-\iid-ness by controlling the $\alpha$ values of the Dirichlet distribution. The smaller $\alpha$ is, the less similar are the clients' local data distributions. The client number is set to $K=6$ as it assures that some of the clients will hold only one data category when the non-\iid-ness is extreme (\Figref{fig:distribution}, $\alpha=0.01$, client No.4, 5). The top row of \Figref{fig:distribution} showcases the non-\iid distribution of the CIFAR10 dataset~\cite{krizhevsky2009learning} over 6 clients.
The setting of SimCLR remains the same in all the experiments, i.e., backbone network set as ResNet18~\cite{he2016deep}, temperature $\tau=0.4$, batch size $B=1024$, learning rate $\eta=1e-3$ and is optimized for 200 epochs using Adam~\cite{kingma2014adam}. 
The supervised method is locally trained for 100 epochs with the learning rate $\eta=1e-3$ and batch size $B=128$, optimized by the Adam optimizer. 
To achieve a fair comparison, we evaluate the linear probing accuracy~\cite{zhang2016colorful} for both SimCLR and supervised method, i.e., we fix the backbone network and retrain a linear classifier on top of it. The accuracies are reported in \Figref{fig:distribution}.

\begin{figure*}[t]
  \vspace{-.8em}
  \centering
  \includegraphics[width=1\textwidth]{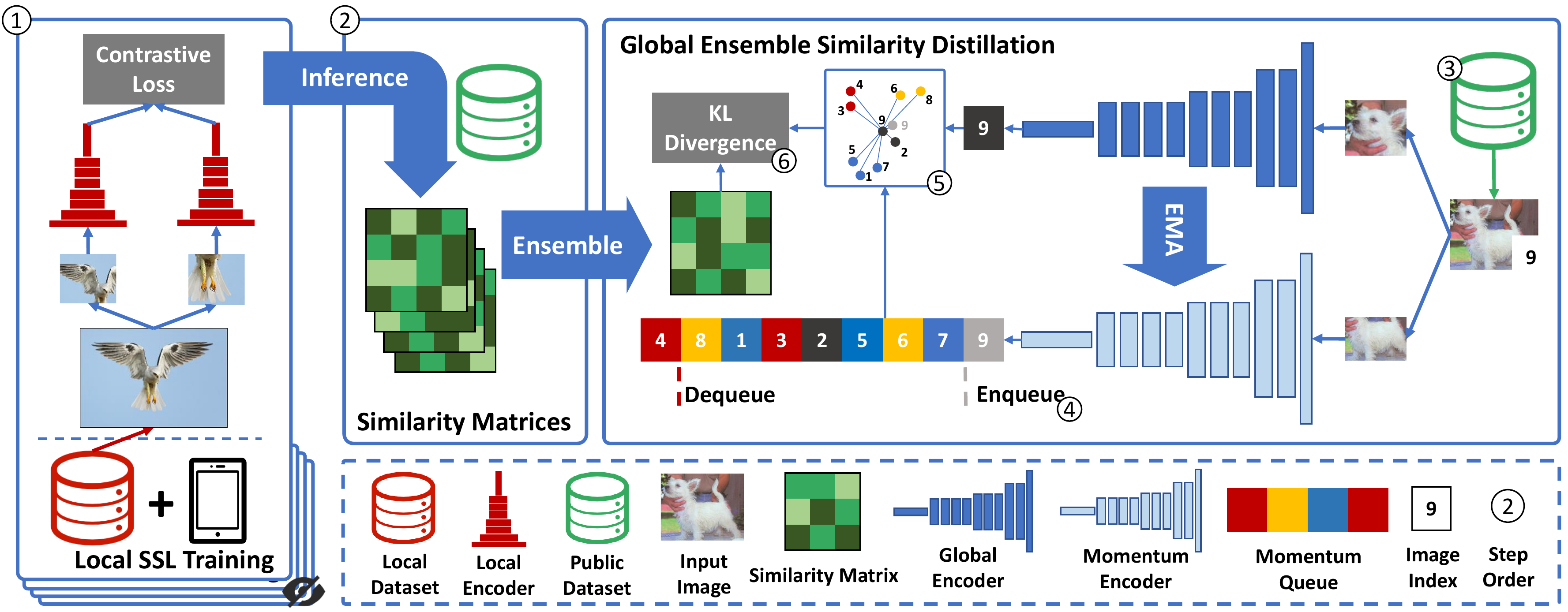}
  \vspace{-1em}
  \caption{Federated representation Learning via Ensemble Similarity Distillation~(\ours). There are three main steps during single communication round: (i) local SSL training, (ii) similarity matrix ensemble, and (iii) global ensemble similarity distillation. The specific steps are broken down and annotated in the figure:
  \ding{192}-local self-supervised contrastive learning; 
  \ding{193}-similarity matrix inference and ensemble on the public dataset; 
  \ding{194}-global aggregation on the public dataset;
  \ding{195}-momentum queue update; 
  \ding{196}-similarity distribution calculation;
  \ding{197}-KL divergence loss calculation and model update.
  }
  \label{fig:overall}
  \vspace{-0em}
\end{figure*}

\subsubsection{Analysis.} 
As the non-\iid-ness increases from $\alpha=100$ to $\alpha=0.01$, the average accuracy of the supervised method decreases from 73.0\% to 18.5\% (-54.5\%). The linear probing accuracy also drops drastically: from 73.5\% to 38.6\% (-34.9\%). However, the linear probing accuracy of SimCLR remains with a subtle performance drop: from 72.8\% to 66.7\% (-6.1\%). Even when the client holds only one single data category, which causes the supervised model to fail, SimCLR's performance maintains at an acceptable level (60.4\% \& 62.4\% for client No.4 \& No.5, respectively). This phenomenon indicates that self-supervised contrastive learning is more robust to the non-\iid-ness of the client data.

We conjecture the main cause of the observed robustness is that the contrastive learning paradigm learns the invariability created by the random data augmentations, which is not severely influenced by lacking certain categories of data and their corresponding annotations. Why and under what conditions self-supervised learning methods such as SimCLR are more robust to the non-\iid data distribution is an intriguing question of its own but is beyond the scope of this paper. Nevertheless, such robustness is exploitable in federated learning since now we can aggregate the local models with an even lower frequency than federated supervised learning: the supervised counterpart often has a comparatively high communication frequency to prevent meaningless local training~\cite{konecny_federated_2016, li2020federated, lin2020ensemble}. 

\section{Federated Representation Learning via Ensemble Similarity Distillation}\label{sec:method}

\subsection{Preliminary: Federated Learning}
Federated learning~(FL) studies how to learn from the data examples distributed across different clients in an extensive network, with all the clients' data privacy protected. Specifically, we seek to minimize the objective:
\begin{align}
    \min_{w} L(w) = \sum_{k=1}^K p_k L_k(w) = \E_k\left[ L_k(w) \right],
\end{align}
where $K$ is the number of the clients, $p_k \geq 0$ representing the importance of the $k$-th client, and $\sum_k p_k = 1$. Due to the intrinsic systematical and statistical heterogeneity of the varying clients, each client might have different data distribution $\gD_k$ and training objective $l_k$: 
\begin{align}
    L_k(w) \triangleq \E_{x \sim \gD_k} \left[ l_k(x; w) \right].
\end{align}

Unlike traditional distributed training that synchronizes the model at each local update, federated learning is often faced with the problem of network thrashing and the varying speed of local training~\cite{konecny_federated_2016, zhao2018federated, li2020federated, lin2020ensemble}. It thus cannot be optimized in a communication-frequent manner. A commonly adopted measure is to train each client on the local data for certain epochs and then communicate their results for model aggregation. For simplicity, this paper mainly focuses on training the local model over a unified self-supervised contrastive objective. However, our proposed method is not limited to such scenario and is applicable when varying local objectives are adopted.

\subsection{Local Self-Supervised Training}
Many federated systems are composed of clients that cannot provide annotations on the local data in the real world. This section describes the existing self-supervised contrastive learning that is adopted as the local training objective for the rest of the paper. Contrastive learning is motivated by the ``multi-view hypothesis'', that the different views of the same data example are likely to represent a close or even the same semantics. Following \citet{oord2018representation}, the objective $\lcl$ trains the model to discern one data point's augmented view (positive) from the rest of the data~(negative):
\begin{align}
    \lcl(f) \triangleq - \E \left[ \log \frac{e^{f(x)^{\top}f(x^+)/\tau}}{e^{f(x)^{\top}f(x^+)/\tau} + \sum_{m=1}^{M_\textnormal{neg}} e^{f(x)^{\top}f(x^-_m) / \tau}}\right], \label{eq:cl}
\end{align}
where $f$ is the encoder network; $x^+$ is the positive view created by a series of data augmentations; $x^-$ is the negative examples independently sampled from the data distribution, and $M_{\textnormal{neg}}$ represents the negative sample size. Typically, in the framework of SimCLR, the negative samples are collected from the other samples within the same batch $M_{\textnormal{neg}}=B-1$. As mentioned above, this training objective is applied across all the local devices, i.e., $l_k = \lcl, \forall k \in \{1, \cdots, K\}$.

In some cases where the non-\iid-ness is extreme, the local client contains only one category of the data, and the negative samples all become falsely negative. Although it leads to a certain extent of the performance decrease (as verified in \Secref{sec:obs}), the contrastive learning objective remains to be conceptually working: it aims to discern subtle differences among the instances regardless of their semantic labels.

\subsection{Global Ensemble Similarity Distillation~(ESD)}
\subsubsection{Similarity matrix ensemble.} 
To enable federated training with the model heterogeneity and pose strict constraints on the clients' data privacy, we assume there is a publicly accessible dataset $\Dpub$ for the global model aggregation. Specifically, the local knowledge of the clients is extracted through the inferred similarity matrices on the public dataset.

Suppose the public dataset $|\Dpub|=N$ has $N$ examples. The $k$-th client's representation matrix $R_k$ is defined as $R_k \triangleq f_k(\Dpub) \in \mathbb{R}^{d_k\times N}$, where $f_k$ is the encoder network; the representation yielded by $f_k$ is of $d_k$ dimension and is automatically normalized to unit-length. 
The similarity matrix $M_k$ of $k$-th client is computed as:
\begin{align}
    M_k = R_k^{\top} R_k,
\end{align}
where $M_{k, ij}=M_{k, ji}=\langle R_{k,:i} , R_{k,:j} \rangle$ is the similarity of the $i$-th and $j$-th element. 

At each communication round, \ours collects the similarity matrices from a fraction $C$ of the clients, denoted as $S=\left\{ M_k \right\}$, the cardinality of which is $|S|=CK$. Then it ensembles the similarity matrices as follows:
\begin{align}
    \hat{M}_{k,ij} &= \exp\left({M_{k, ij}/\tau_{T}}\right), \forall k,i,j \\
    M &= \frac{1}{|S|} \sum_{k=1} \hat{M}_k, \label{eq:sim-ensemble}
\end{align}
where $\tau_T$ is the target temperature regulating the sharpness of each local similarity matrix before averaging. The smaller $\tau_T$ is, the more spike is the similarity matrix, and the model pays more attention to the representations in a small neighborhood. In practice, $\tau_T$ needs to be smaller than 1 to achieve a good performance. Strictly speaking, the ensembled matrix $M$ yielded by \Eqref{eq:sim-ensemble} no longer remains a similarity matrix due to the change of the value scope, $M_{ij} \in (0, +\infty)$. However, as shown in the following section, this design is in accordance with the similarity-based knowledge distillation technique and will not be a problem. 

\begin{algorithm}[h]
    \SetAlgoLined
    \SetArgSty{textnormal}
    \setcounter{AlgoLine}{0}
    \SetKwFunction{server}{\textsc{Server}}
    \SetKwFunction{client}{\textsc{ClientUpdate}}
    \SetKwProg{proc}{Procedure}{}{}

    \proc{\server{}}{
        $w^{(0)} \leftarrow $ random weight initialization \;
    
        \For{each communication round $t=1,\cdots, T$}{
            $S^{(t)} \leftarrow $ random subset ($C$ fraction of the $K$ clients) \;
            \For{each client $k \in S^{(t)}$ \bf{in parallel}}{
                $M^{(t)}_k \leftarrow$ \client{$k$, $w^{(t-1)}$}
            }
            $M^{(t)} \leftarrow$ ensemble from  $\{M^{(t)}_k\}_{k=1}$ \;
            
            $w^{(t)}_0 \leftarrow w^{(t-1)}$ \;
            
            \For{each iteration $j=1,\cdots, J$}{
                sample a mini-batch of data from $\Dpub$ \;
                 
                update $w^{(t)}_j$ by $M^{(t)}$ through \Eqref{eq:esd} \;
            }
            
            $w^{(t)} \leftarrow w^{(t)}_J$
        }
        \KwRet $w^{(T)}$\;
    }\;
    
    \proc{\client{k, w}}{
        $w^{(0)}_0 \leftarrow w^{(0)} \leftarrow w$ \;
        
        \For{each local epoch $e=1,\cdots, E$}{
            \For{each training round $j=1,\cdots, J$}{
                sample a mini-batch of data from $D_k$ \;
                
                update to $w^{(e)}_j$ through \Eqref{eq:cl} \;
            }
            
            $w^{(e)} \longleftarrow w^{(t)}_J$
        }
        $M_k \longleftarrow$ similarity inference on $\Dpub$ with $w^{(E)}$ \;
        
        \KwRet $M_k$\;
    }
  \caption{\textbf{F}ederated representation \textbf{L}earning via \textbf{E}nsemble \textbf{S}imilarity \textbf{D}istillation (\textbf{\ours}). }
  \label{algo:flesd}
\end{algorithm}

\begin{table*}[t]
\vspace{-.8em}
\begin{center}
\tablestyle{3.4pt}{1.4}
\begin{tabular}{c|c|c|ccc|ccc|ccc|ccc}

 & \multirow{2}{*}{\shortstack{Privacy-Leakage \\ Risk}} & \multirow{2}{*}{\shortstack{Comm. \\ Rounds $T$}} &
 \multicolumn{3}{c|}{CIFAR10} &  
 \multicolumn{3}{c|}{CIFAR100} & 
 \multicolumn{3}{c|}{Tiny-ImageNet} &
 \multicolumn{3}{c}{ImageNet-100}
 \\
 & & &
\multicolumn{1}{c}{$\alpha$=100} &
\multicolumn{1}{c}{$\alpha$=1} &
\multicolumn{1}{c|}{$\alpha$=0.01} &
\multicolumn{1}{c}{$\alpha$=100} &
\multicolumn{1}{c}{$\alpha$=1} &
\multicolumn{1}{c|}{$\alpha$=0.01} &
\multicolumn{1}{c}{$\alpha$=100} &
\multicolumn{1}{c}{$\alpha$=1} &
\multicolumn{1}{c|}{$\alpha$=0.01} &
\multicolumn{1}{c}{$\alpha$=100} &
\multicolumn{1}{c}{$\alpha$=1} &
\multicolumn{1}{c}{$\alpha$=0.01} 
\\

\shline
Non-FL    & - & $\infty$ & - & 86.3 & - & - & 55.5 & - & - & 42.9 & - & - & 78.0 & - \\
Min-Local & - & 0 & 62.9 & 66.5 & 60.4 & 12.4 & 41.8 & 41.2 & 35.0 & 34.7 & 35.3 & 63.7 & 61.5  & 59.0 \\
\hline
FedAvg    & High & 10 & 75.6 & 75.2 & \textbf{69.9} & 37.8 & 43.2 & \textbf{44.8} & 37.5 & \textbf{37.3} & \textbf{38.4} & \textbf{68.5} & \textbf{69.5} & \textbf{66.6} \\
FedProx   & High & 10 & 71.3 & 72.3 & 67.5 & 33.9 & 39.4 & 40.6 & 25.1 & 26.0 & 26.3 & 46.1 & 44.6 & 43.2 \\
\hline
\ours     & \textbf{Low} & \textbf{2} & \textbf{75.8} & \textbf{76.2} & 69.6 & \textbf{39.2} & \textbf{43.8} & 42.2 & \textbf{38.1} & 37.2 & 36.5 & 66.3 & 65.4 & 58.7 \\
\ours-cc  & \textbf{Low} & \textbf{1} & 73.2 & 74.4 & 66.2 & 18.9 & 42.1 & 41.4 & 36.4 & 36.4 & 36.5 & 63.8 & 64.3 & 60.6 \\
\end{tabular}
\end{center}
\vspace{-0.7em}
\caption{Linear probing accuracy of \ours and other FURL methods, with SimCLR as the local training scheme. The total epoch of local training is set as constant $E_\text{total}$=$T\times E_\text{local}$=200. 
$\alpha$: value of Dirichlet distribution controlling the non-\iid-ness of the clients' data. Smaller $\alpha$ denotes more severe non-\iid-ness.
\textbf{Non-FL}: upper bound performance trained on the full dataset.
\textbf{Min-Local}: lower bound performance of local training with no global aggregation.
\textbf{\ours-cc}: the degenerate form ``constant communication'' of \ours.
Besides achieving comparable performances as the weight-averaging baselines, \textbf{\ours requires fewer rounds of communication and has strict data privacy protection.} 
}
\label{tab:main}
\vspace{0em}
\end{table*}

\subsubsection{Similarity-based distillation.} 
Unlike traditional knowledge distillation techniques that minimize the cross-entropy between the teacher and student networks' softened class probabilities, similarity-based distillation aims to train a student network that mimics the way the teacher network distributes a population of the representations in the feature space. Our goal is to leverage the ensemble similarity matrix as an off-line teacher network guiding the training of the student network. 

Suppose a series of anchor images $\{x_{j_1}, \cdots, x_{j_m}\}$ are indexed by $\{j_1, \cdots, j_m\}$ where $m$ is the size of the anchor set. For a given query image $x_i$ and a student network $f_S$, the similarity between $x_i$ and the anchor images is computed as $\{ f_S(x_{j_1})^{\top} f_S(x_i), \cdots, f_S(x_{j_m})^{\top} f_S(x_i) \}$. To note here that each representation $f_S(\cdot)$ is also automatically normalized to the unit-length vector as discussed before. Furthermore, we can define a probability distribution over the anchor images based on the similarity scores computed. For the student network, the probability of the $i$-th query on the $j$-th anchor is:
\begin{align}
     q_{j}^{i}=\frac{\exp \left( f_S(x_{j})^{\top} f_S(x_i) / \tau_S\right)}{\sum_{u=1}^{m} \exp \left( f_S(x_{j_u})^{\top} f_S(x_i) / \tau_S\right)}, 
\end{align}
where $\tau_S$ is the student temperature hyper parameter. We set $\tau_S = \tau_T$ following the convention of CompRess~\cite{koohpayegani2020compress} because this restriction assures that the probabilities of the student network and target are roughly same-scaled and have close semantic meaning. Different from CompRess and SEED~\cite{fang2021seed} that have access to the teacher model, we can only perform the distillation on the similarity matrices for the sake of data privacy. The target probability $p^i$ is retrieved from the ensembled similarity matrix $M$, where the i-th query's probability over the j-th anchor is:
\begin{align}
     p_{j}^{i}=\frac{M_{ij}}{\sum_{k=1}^{m} M_{ij_k}}.
\end{align}

Finally, the global aggregation training objective $L_\textnormal{g}$ is the mean KL-divergence between the probabilities over the anchor images of the target and the student network:
\begin{align}
     L_\textnormal{g}(f_S) \triangleq \frac{1}{|\Dpub|} \sum_{i=1} \operatorname{KL}\left( p^i \| q^i \right).
     \label{eq:esd}
\end{align}

During the ensemble similarity distillation, the student network is constantly updated. Therefore re-computing the representations of the anchor images is required, which brings large computational overhead. To address this problem, we follow \citet{he2020momentum} and adopt the momentum encoder and momentum queue to maintain the recently computed representations of the anchor images. The momentum encoder is the slow version of the student network and is updated by exponential moving average~(EMA). 
Suppose the student network $f_\theta$is parameterized by $\theta$ and the momentum encoder $f_\mu$ is parameterized by $\mu$. At each iteration $t$ during ESD, we perform the following:
\begin{align}
    \mu^{(t+1)} = \zeta \cdot \mu^{(t)} + (1-\zeta)\cdot\theta^{(t)}, 
\end{align}
where $\zeta$ is the momentum encoder factor $\zeta$ controlling how slow the momentum encoder is updated towards the student network. When $\zeta=0$, it degenerates to the case where no momentum encoder is adopted.
At each step, the momentum encoder will first encode the mini-batch of the data and update the momentum queue in a first-in-first-out~(FIFO) manner. This way, we spare the computational cost of forward passing the anchor images and achieve better training efficiency.
We present the full algorithmic description of \ours in \Algref{algo:flesd}.
\section{Empirical Experiments}\label{sec:exp}
\subsection{Setup}
\subsubsection{Datasets and models.} We benchmark our proposed framework \ours, along with the baselines on four visual datasets, including CIFAR10/CIFAR100~\cite{krizhevsky2009learning}, Tiny-ImageNet~\cite{le2015tiny}, and ImageNet-100~\cite{tian2019contrastive, wang2020understanding}. The first two datasets are composed of 32x32 RGB images; Tiny-ImageNet's data is of size 64x64; and ImageNet-100 is 100-class subset of ImageNet and retains the original image quality. To simulate different levels of the non-\iid data distribution, we set the $\alpha$ value of the Dirichlet distribution to $\alpha=100, 1, 0.01$ and we benchmark all the methods on them. We use ResNet18 as the encoder network for CIFAR10/CIFAR100/Tiny-ImageNet and ResNet50 for ImageNet-100~\cite{he2016deep}. 

\subsubsection{Evaluation metric.} We adopt the linear probing accuracy~(also known as linear evaluation protocol) to evaluate the representation space quality yielded by the federated self-supervised learning frameworks~\cite{zhang2016colorful, chen2020simple, he2020momentum}. Specifically, after training, we fix the backbone network parameters and retrain a linear classifier on top of it.

\subsubsection{Local self-supervised training.} Same local training strategy is applied to \ours and all the baseline methods: trained by SimCLR framework with learning rate $\eta$=1e-3 and Adam optimizer. Batch size, temperature, and data augmentation are set differently for varying datasets for better performance. Due to the space limitation, we summarize the hyperparamters of local training in supplementary materials.

\begin{figure*}[t]
    \vspace{0em}
    \centering
    \includegraphics[width=1\textwidth]{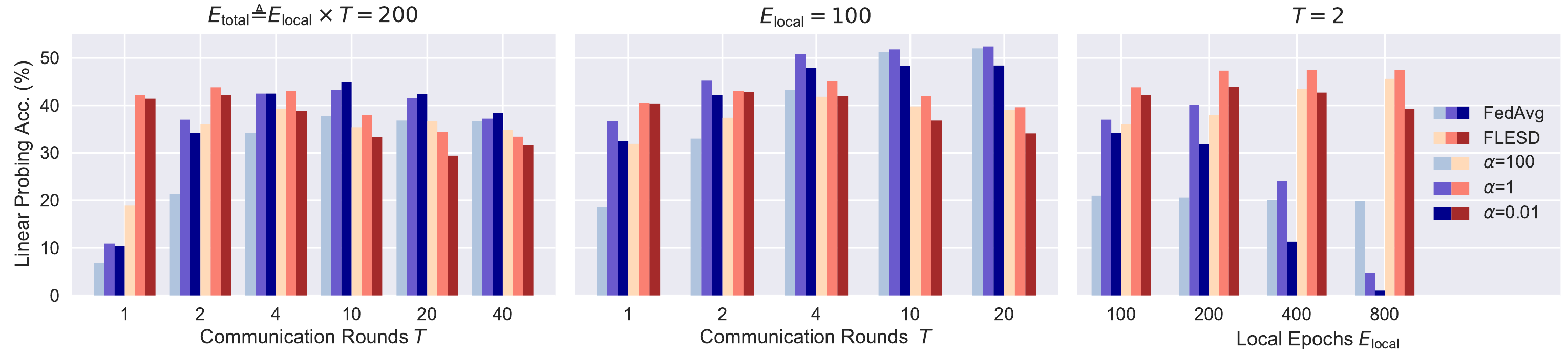}
    \vspace{-1.3em}
    \caption{Influence of different communication schemes, evaluated on CIFAR100. 
    \textbf{Left}: fixed total epochs $E_\text{total}$=200. The performance peak of \ours ($T$=2) appears earlier than FedAvg's ($T$=10) in terms of communication rounds.
    \textbf{Middle}: fixed local epochs $E_\text{local}$=100. \ours's performance saturates as communication round increases.
    \textbf{Right}: fixed communication rounds $T$=2. When the communication is extremely limited, \ours achieves better performance by increasing local training epochs.
    }
    \label{fig:communication}
    \vspace{-0em}
\end{figure*}

\subsubsection{Baselines.} \ours is designed as the first federated self-supervised learning framework that scales to the model heterogeneity. We compare its performance with the classic weight-averaging-based federated methods, including FedAvg~\cite{konecny_federated_2016} and FedProx~\cite{li2020federated}. We report the best performances of FedAvg in the \Tabref{tab:main} by fixing the total update epochs $E_\textnormal{total}\triangleq T\times E_\textnormal{local}=200$ and grid-searching $T\in \{1,2,4,10,20,40\}$. Then we train FedProx under the FedAvg's optimal setting as described above with grid-searching $\mu\in \{0.001, 0.01, 0.1, 1\}$. Both FedAvg and FedProx's client sampling fraction are set as $C=1.0$.

\subsubsection{\ours global aggregation.}
In terms of global aggregation with ensemble similarity distillation, we adopt the same set of hyperparameters for all four datasets, i.e., learning rate $\eta=1e-3$, Adam optimizer, batch size $B^\prime=128$, epoch $E=200$, temperature of the target and the student network $\tau_T=\tau_S=0.1$, factor of the momentum encoder $\zeta=0.999$, anchor set size $m=2048$. The augmentation scheme during the ensemble distillation phase is set the same as the corresponding augmentation used during local training. \Tabref{tab:main} reports the best result of \ours by grid-searching the communication rounds $T\in \{1,2,4,10,20,40\}$. For a fair comparison, client No.0's data is adopted as the public dataset for the global ensemble similarity distillation, and will not be used during local training.  Other federated counterparts such as FedAvg treat it as a simple client on which local training is performed. The client sampling fraction is also set to $C=1.0$.

\subsection{Comparison with Other Federated Baselines}
\Tabref{tab:main} compares the proposed method with other federated baselines on four datasets. In general, \ours and FedAvg both have higher linear probing accuracy than the local training lower bound on all four datasets, proving the necessity of communication and global aggregation. \ours's performance is comparable to the federated baseline FedAvg on multiple datasets, on some of which is even better. It showcases the effectiveness of our method. What's more, \ours achieves this with significantly fewer communication round ($T$=2 vs $T$=10), demonstrating its communication-efficient property. However, the gap between the federated learning methods and the upper bound performance which is quite large: on CIFAR10/CIFAR100/ImageNet-100, the gaps are over 10 points, which needs to be addressed in the future.

\subsection{Communication Schemes}
For completeness, we study three different types of the communication schemes as in \Figref{fig:communication}. Firstly, when the total epochs is set as constant $E_\text{total}=200$ (left subfigure of \Figref{fig:communication}), we iterate through the set of communication rounds and compare the difference between \ours and the FedAvg baseline. Both methods' performance curve is of a reversed U-shape. The performance peak of \ours appears earlier than FedAvg, which showcases the communication-efficiency of our proposed framework. Secondly, we fix the local training epochs as $E_\text{local}=100$ and allow the model to conduct more communication rounds up to 20 (middle subfigure of \Figref{fig:communication}). There are two takeaways: (i) though the local training epochs in federated supervised learning are normally set to 5-10, FedAvg performs surprisingly well when trained longer on the local clients. This phenomenon can be partially explained by the observed robustness in \Secref{sec:obs}; (ii) \ours's performance quickly saturates as the communication rounds increases, which implies possible future improvements. Thirdly, we consider the case where the learning system's communication is strictly limited ($T$=2), \ours's performance is improved as the local training epochs increase. However, except the case when the local data is \iid ($\alpha=100$), FedAvg gradually fails as the local training epochs increases. This phenomenon shows that our method is improved by better-quality of local training, and disputes potential opposition that \ours simply trains self-supervised model on the public dataset. 
\subsection{Convergence}
\begin{figure}[t]
    \vspace{-0.4em}
    \centering
    \includegraphics[width=0.47\textwidth]{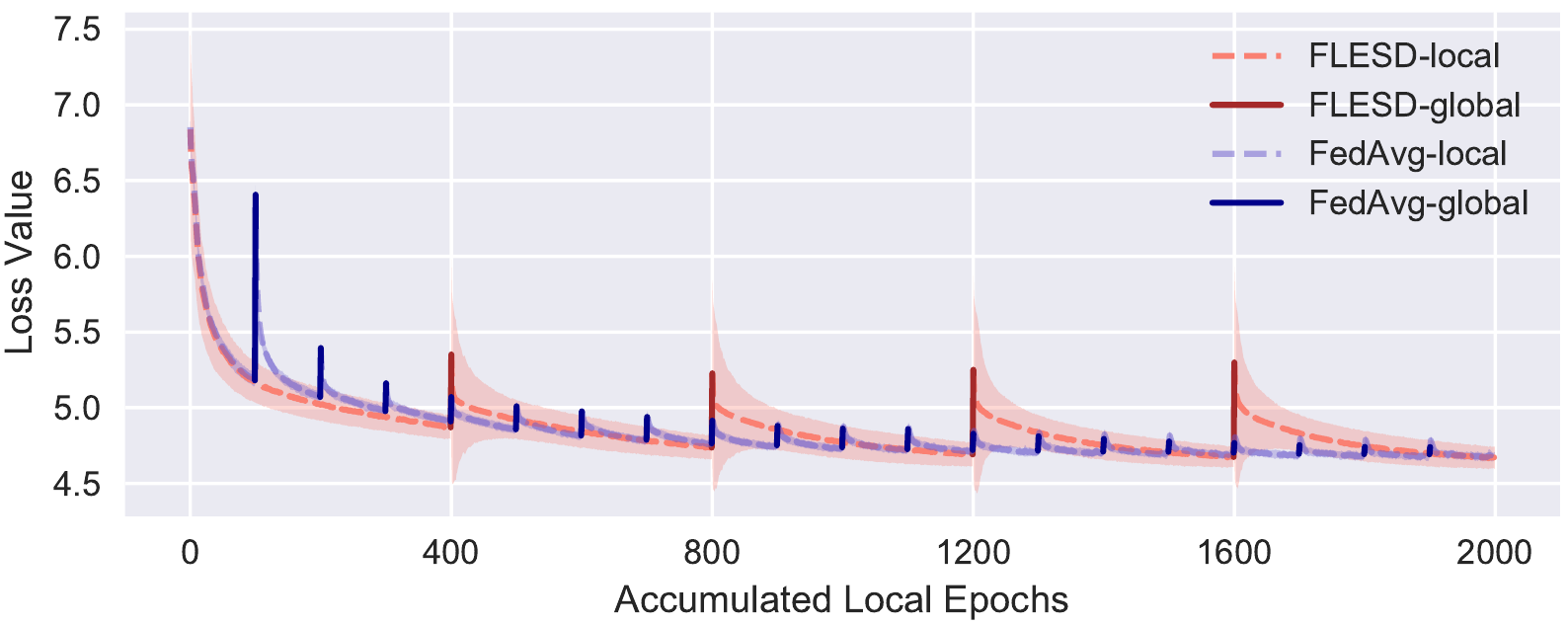}
    \vspace{-1.4em}
    \caption{Training loss trajectories, evaluated on CIFAR100, $\alpha$=1. Dashed lines denote the clients' averaged local training losses. Solid lines denote global aggregation.}
    \label{fig:convergence}
    \vspace{-0em}
\end{figure}

We train \ours and the baseline method FedAvg for longer total epochs $E_\text{total}$=2000 to understand its convergence. Due to the fact that \ours favors fewer communication rounds while FedAvg favors more, here we set $T$=5 and $T$=20 respectively. The self-supervised training loss is reported in \Figref{fig:convergence}. The overall tendency of \ours's training loss is similar to FedAvg. However, after each round of global aggregation, the model is a bit diverged from the optimal point of the self-supervised contrastive loss on the clients. This phenomenon indicates that the model of \ours might have a looser bound on the self-supervised training object, which needs to be further studied and improved.

\section{Discussion of \ours}
\subsubsection{Why similarity matrices of the public dataset?}
One alternative is to send back all the representations yielded by the clients. However, since the clients have misaligned representation spaces after local training, it's more sensible to return their geometric information. 
Besides, similarity matrices can be quantized for lower communication overhead. Secondly, the similarity matrix can be inferred by model of any architecture, which scales \ours to model-heterogeneous systems. Thirdly, inferring on the public dataset does not have privacy concerns, which strictly protects the clients' privacy. 

\subsubsection{On what condition is \ours more communication-efficient than its weight-averaging counterparts?} There two factors to be considered. 
Firstly, suppose a client possesses an encoder network $f_k$, whose parameter number is denoted as $f_k$, communication frequency of \ours and weight-averaging method as $\omega$ and $\omega^\prime$. Then \ours has lower communication cost as long as the following condition holds:
\begin{align}
\omega \cdot |\Dpub| \cdot d_k \leq \omega^\prime \cdot |f_k|,
\end{align}
where we transmit the representation $R_k$ instead of the similarity matrix $M_k$ due to higher communication efficiency. 
Typically, for a public dataset consisting of 10,000 samples and a ResNet18 network which generates 512-dimension feature vectors, $|\Dpub| \cdot d_k = 5.12\text{M} < |f_k| = 11\text{M}$, let alone when the communication frequency is taken into consideration: $\omega < \omega^\prime$. 

The second issue is the transmission and storage of the public data $\Dpub$. Ideally, if a client device has spare storage space for the $\Dpub$, then at the beginning of the whole training process, the server will send $\Dpub$ once and for all, which yields constant communication cost. Otherwise, the client will have to require $\Dpub$ at every global aggregation phase, bringing more communication overhead. However, this issue could also be alleviated by more intricate data transmission scheme, for example, peer-to-peer method. 

\subsubsection{Comparison with existing techniques.} The most significant difference between \ours and other distillation-based FL frameworks including FedDF and FedAD~\cite{lin2020ensemble,gong2021ensemble} is that their clients' output are all naturally aligned towards the same downstream task such as classification. Our proposed framework is dedicated to solve the problem of misaligned representation space for the general federated representation learning. 

Another related research topic is knowledge amalgamation~(KA), which aggregates multiple off-the-shelf teacher models' knowledge by representation-level alignment techniques~\cite{luo2019knowledge,ye2019student,shen2019amalgamating}. However, most of the KA frameworks consider no more than three teacher models and they are not constrained under the privacy protection policy, which means their methods can leverage the teacher models' intermediate state and parameters for better aggregation performance. \ours aims to aggregate multiple client models' knowledge under strict privacy constraints, which requires no exposure of the model parameters.

\subsubsection{Future work.} Future work on theoretical and applicational aspects of \ours can be made. Theoretically, since the optimization goal of ESD is not exactly the same as the local self-supervised training, the convergence guarantee of \ours should be further studied: under what specific assumption of the local and public data will \ours perform reasonably well and how can we improve it? Empirically, one natural application of \ours is federated representation learning on mobile device networks. The challenge of such scenario lies in different aspects as discussed in this paper. For example, the imbalanced computational resource between the data center and edge devices requires the practitioners to answer the critical question of whether transferring small models' knowledge to the large model on the data center will boost the performance?


\section{Conclusion}
\label{sec:conclusion}
In this paper, we propose a novel Federated representation Learning framework via Ensemble Similarity Distillation~(\ours). Unlike existing weight-averaging FURL methods, it adopts the technique of Ensemble Similarity Distillation~(ESD) to aggregate the clients' learned knowledge at each communication round. The proposed framework achieves comparable experimental results despite its strict constraints. To the best of our knowledge, \ours is the first method that works in a communication-efficient and privacy-preserving manner. Furthermore, it has potential application in a FL system with model heterogeneity.

\bibliography{aaai22}

\begin{thebibliography}{44}
\providecommand{\natexlab}[1]{#1}

\bibitem[{Chen et~al.(2020{\natexlab{a}})Chen, Kornblith, Norouzi, and
  Hinton}]{chen2020simple}
Chen, T.; Kornblith, S.; Norouzi, M.; and Hinton, G. 2020{\natexlab{a}}.
\newblock A simple framework for contrastive learning of visual
  representations.
\newblock In \emph{International conference on machine learning}, 1597--1607.
  PMLR.

\bibitem[{Chen et~al.(2020{\natexlab{b}})Chen, Kornblith, Swersky, Norouzi, and
  Hinton}]{chen2020big}
Chen, T.; Kornblith, S.; Swersky, K.; Norouzi, M.; and Hinton, G.
  2020{\natexlab{b}}.
\newblock Big self-supervised models are strong semi-supervised learners.
\newblock \emph{arXiv preprint arXiv:2006.10029}.

\bibitem[{Chen et~al.(2020{\natexlab{c}})Chen, Fan, Girshick, and
  He}]{chen2020improved}
Chen, X.; Fan, H.; Girshick, R.; and He, K. 2020{\natexlab{c}}.
\newblock Improved baselines with momentum contrastive learning.
\newblock \emph{arXiv preprint arXiv:2003.04297}.

\bibitem[{Chen and He(2020)}]{chen2020exploring}
Chen, X.; and He, K. 2020.
\newblock Exploring Simple Siamese Representation Learning.
\newblock \emph{arXiv preprint arXiv:2011.10566}.

\bibitem[{Chen, Xie, and He(2021)}]{chen2021empirical}
Chen, X.; Xie, S.; and He, K. 2021.
\newblock An empirical study of training self-supervised vision transformers.
\newblock \emph{arXiv preprint arXiv:2104.02057}.

\bibitem[{Fang et~al.(2021)Fang, Wang, Wang, Zhang, Yang, and
  Liu}]{fang2021seed}
Fang, Z.; Wang, J.; Wang, L.; Zhang, L.; Yang, Y.; and Liu, Z. 2021.
\newblock Seed: Self-supervised distillation for visual representation.
\newblock \emph{arXiv preprint arXiv:2101.04731}.

\bibitem[{Gong et~al.(2021)Gong, Sharma, Karanam, Wu, Chen, Doermann, and
  Innanje}]{gong2021ensemble}
Gong, X.; Sharma, A.; Karanam, S.; Wu, Z.; Chen, T.; Doermann, D.; and Innanje,
  A. 2021.
\newblock Ensemble Attention Distillation for Privacy-Preserving Federated
  Learning.
\newblock In \emph{Proceedings of the IEEE/CVF International Conference on
  Computer Vision}, 15076--15086.

\bibitem[{Grill et~al.(2020)Grill, Strub, Altch{\'e}, Tallec, Richemond,
  Buchatskaya, Doersch, Pires, Guo, Azar et~al.}]{grill2020bootstrap}
Grill, J.-B.; Strub, F.; Altch{\'e}, F.; Tallec, C.; Richemond, P.~H.;
  Buchatskaya, E.; Doersch, C.; Pires, B.~A.; Guo, Z.~D.; Azar, M.~G.; et~al.
  2020.
\newblock Bootstrap your own latent: A new approach to self-supervised
  learning.
\newblock \emph{arXiv preprint arXiv:2006.07733}.

\bibitem[{He et~al.(2020)He, Fan, Wu, Xie, and Girshick}]{he2020momentum}
He, K.; Fan, H.; Wu, Y.; Xie, S.; and Girshick, R. 2020.
\newblock Momentum contrast for unsupervised visual representation learning.
\newblock In \emph{Proceedings of the IEEE/CVF Conference on Computer Vision
  and Pattern Recognition}, 9729--9738.

\bibitem[{He et~al.(2016)He, Zhang, Ren, and Sun}]{he2016deep}
He, K.; Zhang, X.; Ren, S.; and Sun, J. 2016.
\newblock Deep residual learning for image recognition.
\newblock In \emph{Proceedings of the IEEE conference on computer vision and
  pattern recognition}, 770--778.

\bibitem[{Heo et~al.(2019)Heo, Lee, Yun, and Choi}]{heo2019knowledge}
Heo, B.; Lee, M.; Yun, S.; and Choi, J.~Y. 2019.
\newblock Knowledge transfer via distillation of activation boundaries formed
  by hidden neurons.
\newblock In \emph{Proceedings of the AAAI Conference on Artificial
  Intelligence}, volume~33, 3779--3787.

\bibitem[{Hinton, Vinyals, and Dean(2015)}]{hinton2015distilling}
Hinton, G.; Vinyals, O.; and Dean, J. 2015.
\newblock Distilling the knowledge in a neural network.
\newblock \emph{arXiv preprint arXiv:1503.02531}.

\bibitem[{Jin et~al.(2020)Jin, Wei, Liu, and Yang}]{jin2020towards}
Jin, Y.; Wei, X.; Liu, Y.; and Yang, Q. 2020.
\newblock Towards utilizing unlabeled data in federated learning: A survey and
  prospective.
\newblock \emph{arXiv preprint arXiv:2002.11545}.

\bibitem[{Kim, Park, and Kwak(2018)}]{kim2018paraphrasing}
Kim, J.; Park, S.; and Kwak, N. 2018.
\newblock Paraphrasing complex network: Network compression via factor
  transfer.
\newblock \emph{arXiv preprint arXiv:1802.04977}.

\bibitem[{Kingma and Ba(2014)}]{kingma2014adam}
Kingma, D.~P.; and Ba, J. 2014.
\newblock Adam: A method for stochastic optimization.
\newblock \emph{arXiv preprint arXiv:1412.6980}.

\bibitem[{Konecnuy et~al.(2016)Konecnuy, McMahan, Yu, Richtárik, Suresh, and
  Bacon}]{konecny_federated_2016}
Konecnuy, J.; McMahan, H.~B.; Yu, F.~X.; Richtárik, P.; Suresh, A.~T.; and
  Bacon, D. 2016.
\newblock Federated learning: {Strategies} for improving communication
  efficiency.
\newblock \emph{arXiv preprint arXiv:1610.05492}.

\bibitem[{Koohpayegani, Tejankar, and
  Pirsiavash(2020)}]{koohpayegani2020compress}
Koohpayegani, S.~A.; Tejankar, A.; and Pirsiavash, H. 2020.
\newblock Compress: Self-supervised learning by compressing representations.
\newblock \emph{arXiv preprint arXiv:2010.14713}.

\bibitem[{Krizhevsky, Hinton et~al.(2009)}]{krizhevsky2009learning}
Krizhevsky, A.; Hinton, G.; et~al. 2009.
\newblock Learning multiple layers of features from tiny images.

\bibitem[{Lassance et~al.(2020)Lassance, Bontonou, Hacene, Gripon, Tang, and
  Ortega}]{lassance2020deep}
Lassance, C.; Bontonou, M.; Hacene, G.~B.; Gripon, V.; Tang, J.; and Ortega, A.
  2020.
\newblock Deep geometric knowledge distillation with graphs.
\newblock In \emph{ICASSP 2020-2020 IEEE International Conference on Acoustics,
  Speech and Signal Processing (ICASSP)}, 8484--8488. IEEE.

\bibitem[{Le and Yang(2015)}]{le2015tiny}
Le, Y.; and Yang, X. 2015.
\newblock Tiny imagenet visual recognition challenge.
\newblock \emph{CS 231N}, 7(7): 3.

\bibitem[{Li and Wang(2019)}]{li2019fedmd}
Li, D.; and Wang, J. 2019.
\newblock FedMD: Heterogenous Federated Learning via Model Distillation.
\newblock arXiv:1910.03581.

\bibitem[{Li et~al.(2020)Li, Sahu, Zaheer, Sanjabi, Talwalkar, and
  Smith}]{li2020federated}
Li, T.; Sahu, A.~K.; Zaheer, M.; Sanjabi, M.; Talwalkar, A.; and Smith, V.
  2020.
\newblock Federated Optimization in Heterogeneous Networks.
\newblock arXiv:1812.06127.

\bibitem[{Li et~al.(2021)Li, Jiang, Zhang, Kamp, and Dou}]{li2021fedbn}
Li, X.; Jiang, M.; Zhang, X.; Kamp, M.; and Dou, Q. 2021.
\newblock FedBN: Federated Learning on Non-IID Features via Local Batch
  Normalization.
\newblock arXiv:2102.07623.

\bibitem[{Lin et~al.(2020)Lin, Kong, Stich, and Jaggi}]{lin2020ensemble}
Lin, T.; Kong, L.; Stich, S.~U.; and Jaggi, M. 2020.
\newblock Ensemble distillation for robust model fusion in federated learning.
\newblock \emph{arXiv preprint arXiv:2006.07242}.

\bibitem[{Luo et~al.(2019)Luo, Wang, Fang, Hu, Tao, and
  Song}]{luo2019knowledge}
Luo, S.; Wang, X.; Fang, G.; Hu, Y.; Tao, D.; and Song, M. 2019.
\newblock Knowledge Amalgamation from Heterogeneous Networks by Common Feature
  Learning.
\newblock arXiv:1906.10546.

\bibitem[{McMahan et~al.(2017)McMahan, Moore, Ramage, Hampson, and
  y~Arcas}]{mcmahan2017communication}
McMahan, B.; Moore, E.; Ramage, D.; Hampson, S.; and y~Arcas, B.~A. 2017.
\newblock Communication-efficient learning of deep networks from decentralized
  data.
\newblock In \emph{Artificial Intelligence and Statistics}, 1273--1282. PMLR.

\bibitem[{Oord, Li, and Vinyals(2018)}]{oord2018representation}
Oord, A. v.~d.; Li, Y.; and Vinyals, O. 2018.
\newblock Representation learning with contrastive predictive coding.
\newblock \emph{arXiv preprint arXiv:1807.03748}.

\bibitem[{Park et~al.(2019)Park, Kim, Lu, and Cho}]{park2019relational}
Park, W.; Kim, D.; Lu, Y.; and Cho, M. 2019.
\newblock Relational knowledge distillation.
\newblock In \emph{Proceedings of the IEEE/CVF Conference on Computer Vision
  and Pattern Recognition}, 3967--3976.

\bibitem[{Romero et~al.(2014)Romero, Ballas, Kahou, Chassang, Gatta, and
  Bengio}]{romero2014fitnets}
Romero, A.; Ballas, N.; Kahou, S.~E.; Chassang, A.; Gatta, C.; and Bengio, Y.
  2014.
\newblock Fitnets: Hints for thin deep nets.
\newblock \emph{arXiv preprint arXiv:1412.6550}.

\bibitem[{Shen et~al.(2019)Shen, Wang, Song, Sun, and
  Song}]{shen2019amalgamating}
Shen, C.; Wang, X.; Song, J.; Sun, L.; and Song, M. 2019.
\newblock Amalgamating knowledge towards comprehensive classification.
\newblock In \emph{Proceedings of the AAAI Conference on Artificial
  Intelligence}, volume~33, 3068--3075.

\bibitem[{Smith et~al.(2018)Smith, Chiang, Sanjabi, and
  Talwalkar}]{smith2018federated}
Smith, V.; Chiang, C.-K.; Sanjabi, M.; and Talwalkar, A. 2018.
\newblock Federated Multi-Task Learning.
\newblock arXiv:1705.10467.

\bibitem[{Tian, Krishnan, and Isola(2019{\natexlab{a}})}]{tian2019contrastive}
Tian, Y.; Krishnan, D.; and Isola, P. 2019{\natexlab{a}}.
\newblock Contrastive multiview coding.
\newblock \emph{arXiv preprint arXiv:1906.05849}.

\bibitem[{Tian, Krishnan, and Isola(2019{\natexlab{b}})}]{tian2019crd}
Tian, Y.; Krishnan, D.; and Isola, P. 2019{\natexlab{b}}.
\newblock Contrastive representation distillation.
\newblock \emph{arXiv preprint arXiv:1910.10699}.

\bibitem[{van Berlo, Saeed, and Ozcelebi(2020)}]{van2020towards}
van Berlo, B.; Saeed, A.; and Ozcelebi, T. 2020.
\newblock Towards federated unsupervised representation learning.
\newblock In \emph{Proceedings of the Third ACM International Workshop on Edge
  Systems, Analytics and Networking}, 31--36.

\bibitem[{Wang et~al.(2020)Wang, Yurochkin, Sun, Papailiopoulos, and
  Khazaeni}]{wang2020federated}
Wang, H.; Yurochkin, M.; Sun, Y.; Papailiopoulos, D.; and Khazaeni, Y. 2020.
\newblock Federated Learning with Matched Averaging.
\newblock arXiv:2002.06440.

\bibitem[{Wang and Isola(2020)}]{wang2020understanding}
Wang, T.; and Isola, P. 2020.
\newblock Understanding contrastive representation learning through alignment
  and uniformity on the hypersphere.
\newblock In \emph{International Conference on Machine Learning}, 9929--9939.
  PMLR.

\bibitem[{Wu et~al.(2018)Wu, Xiong, Yu, and Lin}]{wu2018unsupervised}
Wu, Z.; Xiong, Y.; Yu, S.; and Lin, D. 2018.
\newblock Unsupervised feature learning via non-parametric instance-level
  discrimination.
\newblock \emph{arXiv preprint arXiv:1805.01978}.

\bibitem[{Ye et~al.(2019)Ye, Ji, Wang, Ou, Tao, and Song}]{ye2019student}
Ye, J.; Ji, Y.; Wang, X.; Ou, K.; Tao, D.; and Song, M. 2019.
\newblock Student becoming the master: Knowledge amalgamation for joint scene
  parsing, depth estimation, and more.
\newblock In \emph{Proceedings of the IEEE/CVF Conference on Computer Vision
  and Pattern Recognition}, 2829--2838.

\bibitem[{Zagoruyko and Komodakis(2016)}]{zagoruyko2016paying}
Zagoruyko, S.; and Komodakis, N. 2016.
\newblock Paying more attention to attention: Improving the performance of
  convolutional neural networks via attention transfer.
\newblock \emph{arXiv preprint arXiv:1612.03928}.

\bibitem[{Zbontar et~al.(2021)Zbontar, Jing, Misra, LeCun, and
  Deny}]{zbontar2021barlow}
Zbontar, J.; Jing, L.; Misra, I.; LeCun, Y.; and Deny, S. 2021.
\newblock Barlow twins: Self-supervised learning via redundancy reduction.
\newblock \emph{arXiv preprint arXiv:2103.03230}.

\bibitem[{Zhang et~al.(2020)Zhang, Kuang, You, Shen, Xiao, Zhang, Wu, Zhuang,
  and Li}]{zhang2020federated}
Zhang, F.; Kuang, K.; You, Z.; Shen, T.; Xiao, J.; Zhang, Y.; Wu, C.; Zhuang,
  Y.; and Li, X. 2020.
\newblock Federated Unsupervised Representation Learning.
\newblock arXiv:2010.08982.

\bibitem[{Zhang, Isola, and Efros(2016)}]{zhang2016colorful}
Zhang, R.; Isola, P.; and Efros, A.~A. 2016.
\newblock Colorful image colorization.
\newblock In \emph{European conference on computer vision}, 649--666. Springer.

\bibitem[{Zhao et~al.(2018)Zhao, Li, Lai, Suda, Civin, and
  Chandra}]{zhao2018federated}
Zhao, Y.; Li, M.; Lai, L.; Suda, N.; Civin, D.; and Chandra, V. 2018.
\newblock Federated learning with non-iid data.
\newblock \emph{arXiv preprint arXiv:1806.00582}.

\bibitem[{Zhuang et~al.(2021)Zhuang, Gan, Wen, Zhang, and
  Yi}]{zhuang2021collaborative}
Zhuang, W.; Gan, X.; Wen, Y.; Zhang, S.; and Yi, S. 2021.
\newblock Collaborative Unsupervised Visual Representation Learning from
  Decentralized Data.
\newblock In \emph{Proceedings of the IEEE/CVF International Conference on
  Computer Vision}, 4912--4921.

\end{thebibliography}

\end{document}